# How LLMs Cite and Why It Matters: A Cross-Model Audit of Reference Fabrication in AI-Assisted Academic Writing and Methods to Detect Phantom Citations

M.Z. Naser, PhD, PE
School of Civil and Environmental Engineering & Earth Sciences, Clemson University, USA
Artificial Intelligence Research Institute for Science and Engineering, Clemson University, USA
E-mail: mznaser@clemson.edu, Website: www.mznaser.com

**Abstract**
Large language models (LLMs) have been noted to fabricate scholarly citations, yet the scope of this behavior across providers, domains, and prompting conditions remains poorly quantified. We present one of the largest citation hallucination audits to date, in which 10 commercially deployed LLMs were prompted across four academic domains, generating 69,557 citation instances verified against three scholarly databases (namely, CrossRef, OpenAlex, and Semantic Scholar). Our results show that the observed hallucination rates span a fivefold range (11.4%–56.8%) and are strongly shaped by model, domain, and prompt framing. Our results also show that no model spontaneously generates citations when unprompted, which seems to establish hallucination as prompt-induced rather than intrinsic. We identify two practical filters: 1) multi-model consensus (≥3 LLMs citing the same work yields 95.6% accuracy, a 5.8× improvement), and 2) within-prompt repetition (≥2 replications yields 88.9% accuracy). In addition, we present findings on generational model tracking, which reveal that improvements are not guaranteed when deploying newer LLMs, and on capacity scaling, which appears to reduce hallucination within model families. Finally, a lightweight classifier trained solely on bibliographic string features is developed to classify hallucinated citations from verified citations, achieving AUC 0.876 in cross-validation and 0.834 in leave-one-model-out generalization (without querying any external database). This classifier offers a pre-screening tool deployable at inference time.

## 1.0 Introduction

Large language models (LLMs) have become embedded in academic writing workflows at a rapid pace (possibly outstripping our understanding of their proper use, merit, and issues). Recent surveys of researchers across disciplines indicate that a growing share of scholars use tools such as ChatGPT, Claude, and Gemini to draft manuscripts, generate literature reviews, and suggest/compile reference lists [1,2]. Despite such apparent convenience, LLMs routinely fabricate references that do not exist via several means (e.g., attaching real author names to fictitious titles, assigning plausible paper titles that were never published, etc.) [3]. Phantom references of this kind are formatted correctly, contextually appropriate, and nearly indistinguishable from legitimate citations without independent verification [4].

As one might expect, there are consequences of undetected fabrication. For example, a phantom citation that survives peer review enters the scholarly record as an implicit declaration that a particular finding exists and supports the argument at hand. Thus, downstream authors who cite the same paper inherit and propagate the fabrication, and create chains of false attribution that are difficult to unwind once established. Recent audits of conference proceedings have found hallucinated references embedded in peer-reviewed submissions that multiple expert reviewers failed to flag, suggesting that the traditional safeguards of scholarly publishing are poorly calibrated for this class of error [5]. The problem is likely to be compounded by the sheer volume of LLM-assisted writing as adoption accelerates.

Further, we currently have no reliable basis for predicting which LLM models, domains, or prompting patterns will produce more or fewer phantom references. The existing literature lacks comparative evidence because many of the studies examine a narrow slice of available models under non-comparable conditions [6,7]. It is also unclear whether the behavior itself has a causal status: whether models fabricate when asked to cite, or whether citation-like strings emerge as a natural byproduct of generation regardless of the prompt. Thus, without answers to these structural questions, neither individual researchers nor institutional policymakers can make informed decisions about when and how to trust, filter, or prohibit LLM-generated citations.

First, prior studies have examined at most 4 models simultaneously, and none have compared current-generation systems from all major commercial providers within a single experimental framework [8–10]. Given the rapid spread of model families (i.e., OpenAI, Anthropic, Meta, DeepSeek, Mistral, and others), the resulting fragmentation makes it difficult to determine whether hallucination rates reflect idiosyncratic model behavior or systematic patterns tied to architecture, training methodology, or developer practices. Second, few to no studies have reported attempts to manipulate the prompting frame to isolate how the temporal framing of a citation request (e.g., asking for "recent" versus "seminal" references) shapes fabrication rates. This distinction is important because researchers drafting literature reviews often seek either recent developments or foundational texts, and these two types may trigger different generation strategies within the model. Third, and most critically, the field lacks an unprompted control condition. Every existing study prompts models to produce citations and then measures fabrication rates, but without establishing whether models spontaneously generate citations in the absence of an explicit request [11]. Thus, it remains unclear whether citation hallucination is an intrinsic tendency or a behavior induced by prompting.

This paper addresses these gaps through extensive experiments. In particular, we evaluate 10 commercially deployed language models across 4 academic domains, 2 temporal framings, 3 independent replications per prompt, and an unprompted control condition. The resulting dataset comprises 69,557 citation instances, of which 40,529 are verified against a three-source resolution pipeline spanning CrossRef, OpenAlex, and Semantic Scholar. Our experimental design produces six contributions that helps bridge the gap within the current evidence base, in which we: 1) quantify hallucination rates across ten models from seven providers, 2) demonstrate that temporal framing substantially modulates fabrication, 3) establish through the unprompted control that no model spontaneously produces formal citations, 4) characterize systematic bibliometric biases in the citations that models do produce, including amplification of open-access works, high-citation papers, and a narrow set of publishers, 5) propose and validate two practical verification heuristics, multi-model consensus and within-prompt repetition, that reduce hallucination rates, and 6) track generational and capacity-scaling trajectories within model families.

The remainder of this paper proceeds as follows. Section 2 reviews prior work on citation fabrication and positions the present study within the broader hallucination evaluation literature.

Section 3 details the experimental design, verification pipeline, and validation methodology. Section 4 presents the primary results, organized by model performance, framing effects, domain effects, and bibliometric bias. Section 5 reports comparative analyses across model taxonomies, including generational, capacity, and openness comparisons, and Section 6 develops the two verification heuristics and evaluates their filtering performance. Section 7 discusses implications, limitations, and directions for future work.

**2.0 Related work**

This section situates the present study within two bodies of literature: empirical investigations of citation fabrication in LLMs and broader efforts to benchmark hallucination across model families.

*2.1 Citation fabrication studies*

The empirical study of LLM citation fabrication began with Walters and Wilder [3], who prompted GPT-3.5 and GPT-4 to generate short literature reviews on 42 multidisciplinary topics and manually verified the resulting 636 references. Their finding that 55% of GPT-3.5 citations and 18% of GPT-4 citations were entirely fabricated established a quantitative baseline and demonstrated that generational improvement within a model family was possible. The same researchers also found that fabricated citations often contained real author names combined with fictitious titles, a pattern that complicates detection through surface-level inspection.

Several studies have since replicated and extended this work along different axes. Linardon et al. [10] examined GPT-4o across mental health topics varying in public familiarity and found that fabrication rates reached 28 to 29% for less visible disorders (e.g., body dysmorphic disorder), compared to 6% for major depressive disorder. This result introduced topic familiarity as a moderating variable and suggested that models are more likely to fabricate when their training data is sparse for a given subject. Mugaanyi et al. [9] pursued a geographic dimension by validating 3,451 citations generated by four models using the CrossRef API. Their analysis showed hallucination rates exceeding 80% in prompts referencing lower-income countries. This particular finding raises concerns about epistemic equity in AI-assisted research tools.

A parallel line of work has examined citation fabrication in specific professional domains. Studies in radiology, psychiatry, and clinical rehabilitation have reported fabrication rates ranging from 20 to 70%, depending on the model version, prompt specificity, and the availability of relevant literature in the model's training corpus [12–14]. In one particular study, Buchanan et al. [15] focused specifically on economics and noted that over 30% of GPT-3.5 citations across Journal of Economic Literature topics were nonexistent, and that reliability decreased as prompts became more specialized.

*2.2 Hallucination benchmarks and detection*

Citation fabrication sits within a broader effort to measure and categorize hallucination in LLMs. From the lens of benchmarking, notable benchmarks such as TruthfulQA [16], TriviaQA [17], and others, evaluate factual accuracy across diverse question types, but none include a citation-specific evaluation component. HalluLens [18], introduced at ACL 2025, proposed a taxonomy distinguishing intrinsic hallucination (inconsistency with training data) from extrinsic hallucination (deviation from input context) and developed dynamic test sets to mitigate data leakage. Under this taxonomy, citation fabrication constitutes a form of extrinsic hallucination in the absence of retrieval augmentation, since the model generates references not grounded in any provided source material [4].

It is worth noting that Anh-Hoang et al. [11] developed an attribution framework that quantifies the relative contribution of prompting strategies vs. model-intrinsic behavior to hallucination. Their Prompt Sensitivity and Model Variability metrics could be used to provide a formal vocabulary for the question that motivates our unprompted control condition. Thus, if citation hallucination is entirely prompt-sensitive, as our results will demonstrate, then mitigation strategies should focus on prompt design and post-hoc verification rather than architectural modification alone. Our verification pipeline draws on the CrossRef, OpenAlex, and Semantic Scholar and applies a tiered confidence scoring system calibrated through automated validation, as described in Section 3.

**3.0 Study design**

This section describes the experimental design, data collection protocol, citation parsing methodology, reference verification pipeline, and the automated validation procedure used to assess pipeline reliability (see Fig. 1). The full prompt templates and scoring rubrics are provided in the Appendix.

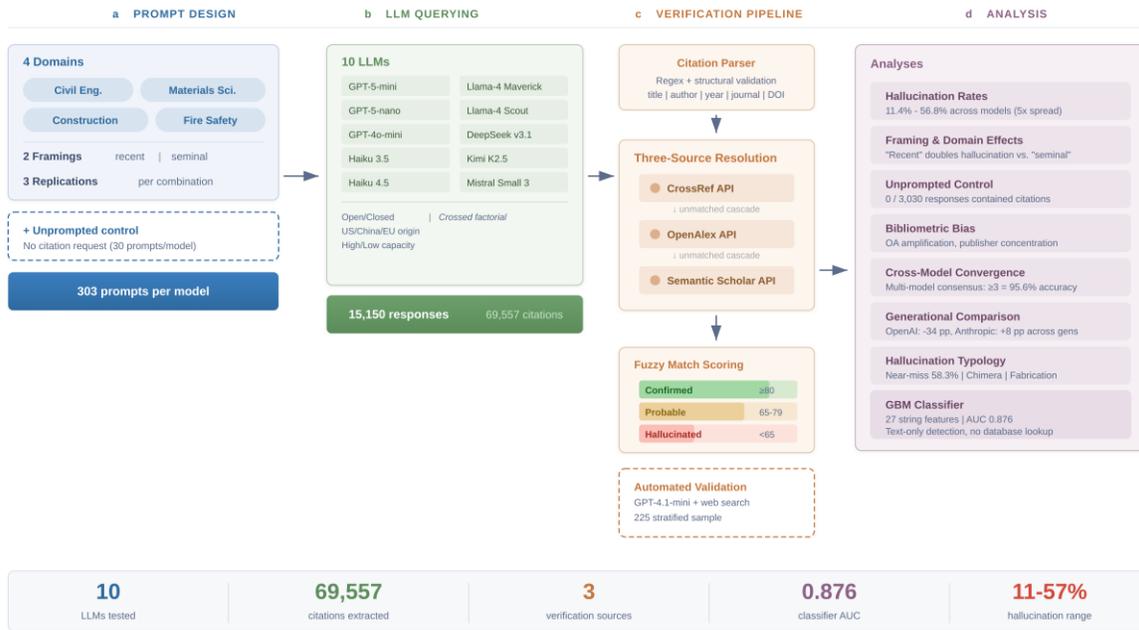

Fig. 1 | Study design and analysis pipeline.
(a) Prompts span 4 engineering domains x 2 framings x 3 replications + unprompted control.
(b) 10 commercially deployed LLMs, balanced across provider, origin, and capacity tiers. (c) Cascading three-source verification with tiered scoring. (d) Eight analysis dimensions.

Fig. 1 Study design

*3.1 Models and taxonomy*

As mentioned earlier, we evaluate 10 LLMs across 7 developers, 3 geographic regions, and both open-weight and closed-weight release strategies. Our selection criteria required that each model: a) be accessible through a public API as of the data collection period, b) support text generation without mandatory retrieval augmentation, and c) represent a distinct point in at least one of three taxonomic dimensions: developer generation, model capacity, or weight accessibility.

This set includes three OpenAI models (GPT-4o-mini, GPT-5-mini, GPT-5-nano), two Anthropic models (haiku-3.5, haiku-4.5), two Meta models (llama4-scout, llama4-maverick), and one model each from DeepSeek (deepseek-v3.1), Moonshot AI (kimi-k2.5), and Mistral (mistral-small-3). This composition enables three structured comparisons. More specifically, generational comparisons hold the provider constant while varying the model version (GPT-4o-mini versus GPT-5-mini, haiku-3.5 versus haiku-4.5). Then, capacity comparisons hold the model family constant while varying the mixture-of-experts configuration or parameter count (llama4-scout versus llama4-maverick, GPT-5-nano versus GPT-5-mini). Finally, openness comparisons group models by whether their weights are publicly released.

### 3.2 Prompt design and data collection

Each model received an identical set of prompts organized in a factorial design crossing four academic domains with two temporal framings, plus an unprompted control condition. The four domains are structural engineering, climate and environmental science, biomedical research, and natural language processing (NLP) and artificial intelligence (AI). These domains were selected to span a range of literature sizes and publication cultures.

For each domain, we constructed prompts asking the model to provide scholarly references on a specific topic. All prompts used identical syntactic structures and varied only the domain-specific content and the temporal frame. The temporal framing variable manipulated whether the prompt requested "recent and influential" references or "seminal and foundational" references. Our design selected this manipulation to target a specific hypothesis: models may hallucinate at different rates when asked to produce references from different temporal windows, because recent publications are less likely to appear in the training data, while seminal works are more widely cited and cross-referenced across multiple documents in the pre-training data.

The unprompted condition presented the same topical questions but did not request citations or references. This condition serves as the control, building on the idea that, if models produce formal citations even when not asked, then some portion of hallucination reflects an intrinsic tendency toward citation generation. If they do not, then all observed fabrication can be attributable to the citation-request framing.

Each prompted condition was repeated three times per model to assess within-model consistency. The complete design yields 303 prompts per model (4 domains × 2 framings × 3 replications for the parametric condition, plus additional candidate-pool prompts and 303 unprompted controls), producing 15,150 total responses across the ten models. Data collection was performed via each model's API.

### 3.3 Citation parsing

We developed a two-stage parsing pipeline to extract structured citation data from model responses. The first stage applies regular expressions to identify reference-like strings, including parenthetical author-year patterns, numbered reference lists, and inline DOI or URL mentions. Then, the second stage normalizes the extracted strings into a canonical form consisting of parsed author names, title, journal or venue, year, and DOI, where present. For the unprompted condition, we deployed a conservative parser with stricter matching criteria to avoid false positives. This parser required at least two of the following: a recognizable author-year format, a journal or conference name, and a DOI or volume-page combination.

The parsing pipeline extracted 69,557 citation instances from the 15,150 responses. The unprompted condition produced zero parseable citations across all ten models and all 3,030 unprompted responses, a result we return to in Section 4.

*3.4 Reference verification pipeline*

Each parsed citation was submitted to a three-source verification pipeline designed to maximize recall while maintaining high precision. The pipeline queries CrossRef, OpenAlex [19], and Semantic Scholar in sequence, and applies fuzzy title matching at each stage. CrossRef serves as the primary authority for DOI resolution and publication metadata. OpenAlex provides broader coverage of open-access and preprint literature, along with bibliometric indicators, while Semantic Scholar adds coverage of computer science and biomedical publications that may be underrepresented in CrossRef, and provides high-quality author disambiguation.

The matching procedure assigns a confidence score on a 0-to-100 scale based on a generic process that accounts for title similarity (weighted at 0.6), author overlap (0.2), and year agreement (0.2). Title similarity is computed using token-level Jaccard distance after lowercasing and removing punctuation, a simple measure that is robust to minor paraphrasing and word-order variation. Author overlap compares the set of parsed last names against the metadata record, accommodating partial matches for multi-author papers, where models may list only the first author or only the first two authors. In our pipeline, citations with a score of 80 or higher on at least one source are classified as "confirmed" matches. Those scoring between 65 and 79 are classified as "probable" matches. All remaining citations are classified as hallucinated.

For confirmed matches, the pipeline retrieves bibliometric metadata from OpenAlex, including the cited-by count, open-access status, publisher, journal name, and publication year. These fields support the bibliometric bias analyses presented in Sections 4 and 5. In total, the pipeline processed 69,557 citations and returned verified matches for 40,529, with a median API response time of 1.2 seconds per citation across the three sources.

*3.5 Automated validation*

To assess the reliability of the verification pipeline, we conducted an independent validation using GPT-4.1-mini with web search enabled [20]. A stratified sample of 225 citations was drawn from three strata: 75 confirmed matches, 75 probable matches, and 75 citations classified as hallucinated. For each citation, the validation model was instructed to search the web for the exact title and authors and to render a judgment of "real," "plausible but unverifiable," or "fabricated."

The validation yielded the following results: among the 75 confirmed matches, all 75 were verified as real (100% precision). Among the 75 probable matches, 71 were verified as genuine (95.2%). Among the 75 hallucinated citations, 8 were found to be real (10.7%), indicating that the pipeline's false-negative rate at the confirmed threshold is low but nonzero. These results support the use of the confirmed threshold (score $\geq$ 80) as the primary analysis target, with the inclusive threshold (score $\geq$ 65) reserved for sensitivity analysis. The 10.7% recovery rate among the hallucinated stratum suggests that the confirmed-only rates reported throughout this paper are conservative estimates of true hallucination.

**4.0 Results**

This section reports the primary empirical findings, organized by model-level hallucination rates, experimental manipulations, bibliometric bias in confirmed citations, cross-model convergence patterns, and sensitivity to the verification threshold. All reported hallucination rates use the confirmed threshold (score ≥ 80) unless otherwise noted. Statistical tests report chi-squared statistics with Cramér's V as the effect size measure, and bootstrap 95% confidence intervals are computed from 10,000 resamples.

*4.1 Hallucination rates by model*

Across the ten models, hallucination rates range from 11.4% (GPT-5-mini) to 56.8% (haiku-4.5), a fivefold variation. Table 1 presents the full model rankings with bootstrap confidence intervals, total citations generated, and the number verified as real.

Table 1 Model-level hallucination rates (and other information)

| Rank | Model | Real | Total | Hallucinated | Halluc. Rate | 95% CI | OA Rate | Median Cites | Openness | Origin |
|---|---|---|---|---|---|---|---|---|---|---|
| 1 | GPT-5-mini | 3,149 | 3,555 | 406 | 11.4% | [10.4, 12.5] | 91.8% | 1,036 | Closed | US |
| 2 | GPT-5-nano | 2,526 | 3,293 | 767 | 23.3% | [21.9, 24.7] | 90.0% | 1,133 | Closed | US |
| 3 | deepseek-v3.1 | 6,023 | 8,471 | 2,448 | 28.9% | [28.0, 29.9] | 85.8% | 668 | Open | Chinese |
| 4 | llama4-maverick | 5,393 | 8,105 | 2,712 | 33.5% | [32.4, 34.5] | 86.5% | 630 | Open | US |
| 5 | GPT-4o-mini | 4,282 | 7,824 | 3,542 | 45.3% | [44.2, 46.4] | 77.5% | 743 | Closed | US |
| 6 | kimi-k2.5 | 3,592 | 6,802 | 3,210 | 47.2% | [46.0, 48.4] | 78.4% | 359 | Open | Chinese |
| 7 | mistral-small-3 | 4,120 | 7,976 | 3,856 | 48.3% | [47.2, 49.4] | 83.1% | 731 | Open | European |
| 8 | haiku-3.5 | 3,964 | 7,745 | 3,781 | 48.8% | [47.7, 49.9] | 82.5% | 743 | Closed | US |
| 9 | llama4-scout | 4,531 | 8,952 | 4,421 | 49.4% | [48.4, 50.4] | 83.9% | 675 | Open | US |
| 10 | haiku-4.5 | 2,949 | 6,834 | 3,885 | 56.8% | [55.6, 58.0] | 77.3% | 362 | Closed | US |

As one can see, the three lowest-hallucinating models (GPT-5-mini at 11.4%, GPT-5-nano at 23.3%, and deepseek-v3.1 at 28.9%) are separated by 17.5% from the remaining seven, which cluster between 33.5 and 56.8%. Further, none of the bootstrap confidence intervals for the top three overlap with those of the bottom seven, which further confirms that this separation is not attributable to sampling variability. A chi-squared test of independence between model identity and hallucination status yields $\chi^2 = 3,842.7$ (degrees of freedom = 9, $p \approx 0$), with Cramér's V = 0.236, indicating a moderate-to-large association between model choice and citation accuracy.

The same table also shows that the two GPT-5 models produced substantially fewer total citations (3,555 and 3,293) than the other eight models, which generated between 5,393 and 9,117 citations each. Since all models received identical prompt sets, this difference reflects variation in citations per response rather than variation in response count.

*4.2 Framing effect*

Prompts requesting "recent and influential" references produced a hallucination rate of 74.1%, compared to 55.0% for prompts requesting "seminal and foundational" references. This 19.1% gap is statistically significant ($\chi^2 = 917.6$, $p < 10^{-201}$), with Cramér's V = 0.115 (a small-to-medium effect consistent across all 10 models). It is interesting to note that every model hallucinated at a higher rate under the recent framing than under the seminal framing, with no exceptions (see Fig. 2). Its magnitude varied from 12% (GPT-5-mini) to 24% (llama4-scout). Such an observation may indicate that, while the direction is universal, the degree of vulnerability to temporal framing with regard to recent years varies across models.

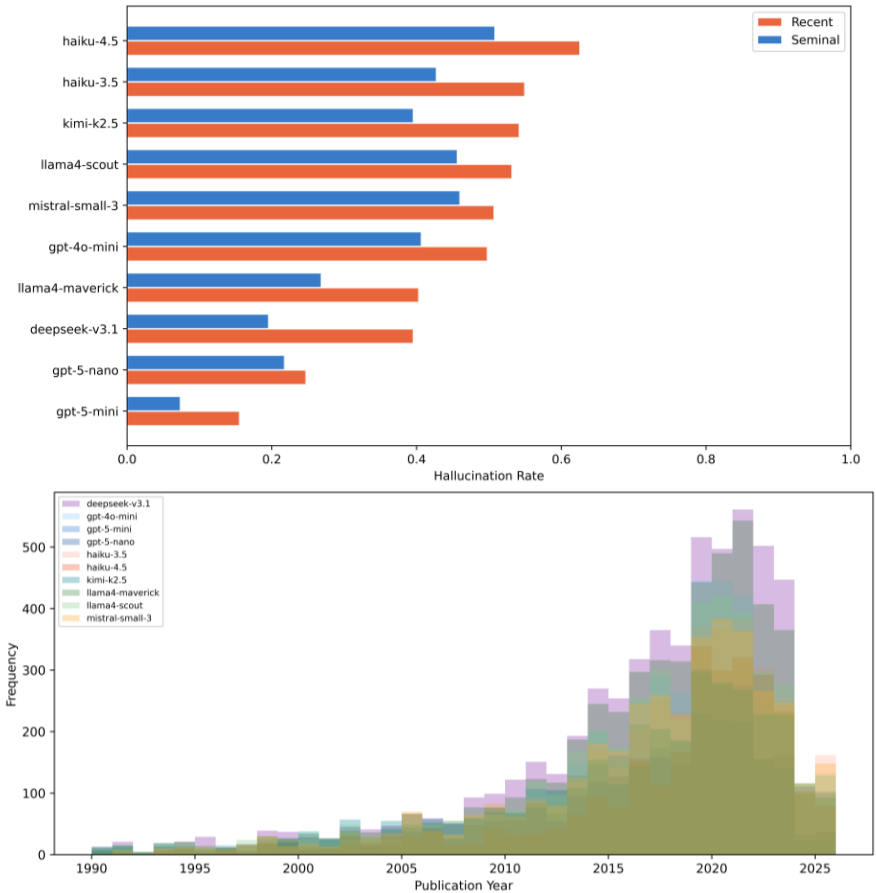

Fig. 2 Framing effect

We believe that the direction of this effect aligns with a training-data-availability account. For example, seminal works are, by definition, frequently cited, widely discussed, and likely to appear in multiple documents within pre-training data. Recent publications, by contrast, may postdate the training data cutoff or appear in too few documents to have been reliably encoded. When prompted to produce recent references, LLMs face a generative task with fewer grounded anchors, and the result is a sharply elevated fabrication rate. This interpretation is consistent with Linardon et al. [10] finding that topic familiarity modulates hallucination, but our result extends the mechanism from domain familiarity to temporal familiarity within the same domain.

*4.3 Domain effect*

Our analysis shows that hallucination rates vary substantially across the four academic domains. NLP and AI exhibited the lowest rate at 26.6%, followed by biomedical research at 39.4%, climate and environmental science at 41.8%, and structural engineering at 50.1%. A chi-squared test confirms this variation ($\chi^2 = 2,603.5$, $p \approx 0$, $V = 0.193$), with domain exerting a larger effect than framing ($V = 0.193$ vs. $V = 0.115$).

This ordering mirrors expected differences in pre-training representation. Overall, NLP and AI research are heavily represented in the arXiv preprints, technical blogs, and conference proceedings that constitute a large share of LLM training data. In contrast, structural engineering publishes primarily through subscription-access journals with smaller readerships and fewer online traces. Between these extremes, biomedical and climate research occupy intermediate positions in terms of both literature volume and open-access penetration. In short, models hallucinate less when the relevant literature is more densely represented in their training data.

*4.4 Bibliometric bias in confirmed citations*
We observed several systematic biases in the verified citations. These appear along three dimensions: open-access status, citation popularity, and publisher concentration.

Global baselines for open-access availability vary by field but typically range from 50 to 55% of the published literature. Table 1 shows that, across all ten models, the share of confirmed citations pointing to open-access works ranges from 77.3% (haiku-4.5) to 91.8% (GPT-5-mini). The same figure also shows that every model in our sample over-represents open-access works by a wide margin, and the degree of over-representation tracks inversely with hallucination rate. GPT-5-mini and GPT-5-nano (at 91.8 and 90.0% OA, respectively) show the strongest amplification. This may suggest that more accurate models draw more heavily from the open-access literature, which is most densely available in the training data.

From this lens, it is worth exploring citation popularity and its role in being selected by LLMs. We report that median cited-by counts for confirmed references range from 359 (kimi-k2.5) to 1,132 (GPT-5-nano), while field-level medians for the four domains in our study fall between 50 and 100 citations. This implies that LLMs do not seem to sample uniformly from their training distributions, but instead they preferentially retrieve highly cited works (which further amplifies the visibility of already-prominent research). This popularity bias is most pronounced in the two GPT-5 models (median cited-by of 1,036 and 1,132, respectively, with amplification statistically significant at $p < 10^{-46}$), and least pronounced in haiku-4.5 and kimi-k2.5, the two models with the highest hallucination rates and the lowest median cited-by counts.

*4.5 Cross-model convergence*
During our experiments, we realized that when multiple models independently produce the same reference in response to the same prompt, the probability that it corresponds to a real publication increases. Our results show that among unique title strings cited by only one model, the match rate against our verification pipeline is 16.5%. For titles cited by two models, this rises to 87.4%. At three or more models, the match rate reaches 95.6%, a 5.8-fold improvement over the single-model baseline. A similar pattern also appears within models across replications. Here, when a model produces the same citation in only one of three replications, the match rate is 28.6%. However, when the same citation recurs in two or more replications, the match rate rises to 88.9% (a 3.1-fold improvement).

The two patterns above point to the same underlying mechanism: hallucinated citations are stochastic and variable, whereas real citations are stable and convergent. This means that a fabricated reference is unlikely to be generated independently by multiple models because it lacks an external anchor in the training data. We measure cross-model agreement using pairwise Jaccard similarity on the sets of unique titles produced by each model to examine this observation further (see Fig. 3). Most model pairs share Jaccard indices between 0.15 and 0.25 (i.e., reflecting modest but nonzero overlap in citation repertoires). This figure also shows that there is higher agreement within-family pairs (e.g., GPT-5-mini and GPT-5-nano share a Jaccard index of 0.540, Llama4-scout and llama4-maverick also show higher agreement (Jaccard = 0.312)). Such an agreement may reflect shared training within the model family.

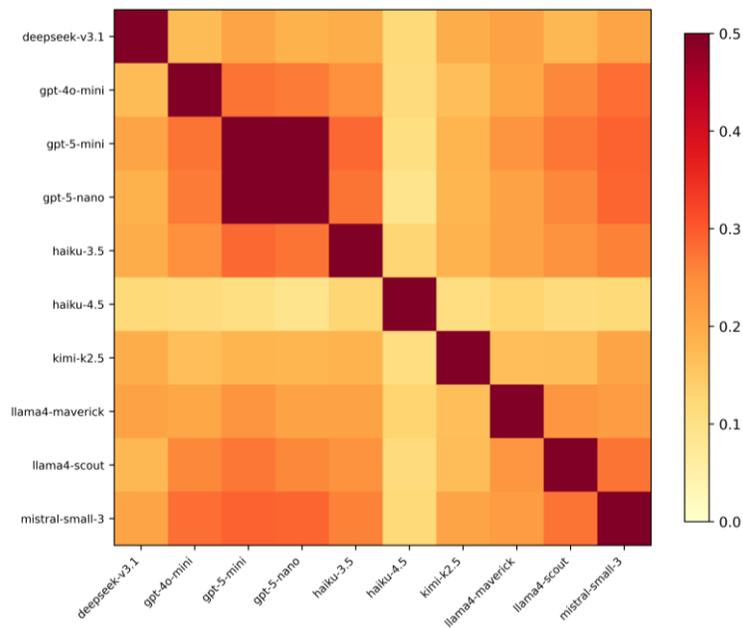

Fig. 3 Cross model citation convergence

It is worth pointing out that haiku-4.5 is a persistent outlier in cross-model agreement, with pairwise Jaccard values ranging from 0.092 to 0.132 (i.e., the lowest in the matrix). This isolation is consistent with haiku-4.5's distinctive publisher distribution, low median citation counts, and high hallucination rate, and it could imply that this model draws on a narrower or differently weighted segment of the scholarly literature than its peers. Whether this reflects intentional design choices or incidental consequences of the training process is not determinable from our data.

*4.6 Sensitivity analysis*

All primary analyses use the confirmed threshold (score ≥ 80). To further assess the stability of our findings under a more inclusive criterion, we recomputed all model-level hallucination rates using the inclusive threshold (score ≥ 65, combining confirmed and probable matches). At this threshold, hallucination rates decrease across all models, as expected, but the rank ordering is largely preserved. Spearman's rank correlation between the confirmed-only and inclusive rankings is ρ = 0.758 (p = 0.011). This indicates partial but significant preservation of the model hierarchy.

Table XXX Rankings post the sensitivity analysis

| Model | Rank (≥80) | Halluc Rate (≥80) | Total (≥80) | Real (≥80) | Rank (≥65) | Halluc Rate (≥65) | Total (≥65) | Real (≥65) |
|---|---|---|---|---|---|---|---|---|
| GPT-5-mini | 1 | 11.4% | 3,555 | 3,149 | 1 | 9.3% | 3,555 | 3,224 |
| GPT-5-nano | 2 | 23.3% | 3,293 | 2,526 | 5 | 17.8% | 3,293 | 2,706 |
| deepseek-v3.1 | 3 | 28.9% | 8,471 | 6,023 | 3 | 14.0% | 8,471 | 7,284 |

| Model | | | | | | | | |
|---|---|---|---|---|---|---|---|---|
| llama4-maverick | 4 | 33.5% | 8,105 | 5,393 | 2 | 11.8% | 8,105 | 7,147 |
| GPT-4o-mini | 5 | 45.3% | 7,824 | 4,282 | 4 | 16.7% | 7,824 | 6,519 |
| kimi-k2.5 | 6 | 47.2% | 6,802 | 3,592 | 10 | 23.8% | 6,802 | 5,186 |
| mistral-small-3 | 7 | 48.3% | 7,976 | 4,120 | 7 | 18.1% | 7,976 | 6,530 |
| haiku-3.5 | 8 | 48.8% | 7,745 | 3,964 | 8 | 19.2% | 7,745 | 6,255 |
| llama4-scout | 9 | 49.4% | 8,952 | 4,531 | 6 | 18.0% | 8,952 | 7,342 |
| haiku-4.5 | 10 | 56.8% | 6,834 | 2,949 | 9 | 22.8% | 6,834 | 5,275 |

That said, two models shifted substantially in relative position under the inclusive threshold. Kimi-k2.5 drops from sixth to tenth, and GPT-5-nano moves from second to fifth. These shifts are likely to reflect differences in the proportion of probable-tier matches (score 65–79) across models (i.e., kimi-k2.5 produces a disproportionate share of citations in this intermediate zone, while mistral-small-3 produces relatively few). For the remaining eight models, rank changes are at most one position. The top three (GPT-5-mini, GPT-5-nano, deepseek-v3.1) and the bottom two (llama4-scout, haiku-4.5) are invariant to threshold choice. This analysis reinforces our confidence that the headline findings are not artifacts of the verification cutoff.

### *4.7 Hallucination typology*

As mentioned earlier, the open literature documents that not all hallucinated citations are fabricated equally. Thus, to characterize the structural diversity of hallucinations, we classified each unverified citation by its best fuzzy match score against the resolution database into four categories: *probable real* (best score 65–79, likely real papers that narrowly missed the confirmation threshold), *near miss* (score 50–64, corrupted versions of real titles), *distant chimera* (score 30–49, loosely inspired by real work but substantially transformed), and *fabrication* (score below 30, no detectable resemblance to any indexed publication).

Across all models, 58.3% of hallucinated citations fell in the probable real category, 31% were near misses, 4.7% were distant chimeras, and 6.0% were outright fabrications. This distribution infers that the dominant hallucination mechanism is not invention but distortion — models retrieve partial bibliographic information and imperfectly reconstruct it to produce outputs that closely resemble but do not exactly match real publications.

The noted typology distribution varied substantially across models. In particular, GPT-5-mini and GPT-5-nano produced distant chimeras at rates of 37 and 29%, respectively, compared to 1–3% for nearly every other model. This qualitative behavioral distinction indicates that OpenAI's newer models hallucinate through a different mechanism (possibly via generating citations loosely inspired by real work rather than producing near-miss corruptions of specific papers). The remaining models clustered around 60% probable real and 30% near miss, consistent with a shared failure mode of partial retrieval and imperfect reconstruction.

### *4.8 Text-based hallucination classifier*

The typological analysis revealed systematic structural differences between verified and hallucinated citations. Thus, we investigated whether these differences are sufficient to classify citation status solely from bibliographic metadata, without querying any external database.

We extracted 27 text-level features from each citation string, spanning five categories: title properties (word count, character count, presence of colons, question marks, numbers, acronyms, capitalization ratio, leading article, template patterns, "toward" prefix), author properties (count, character length, presence of "et al.," initial diversity), year properties (round year, future year, missing flag, recency as 2025 minus publication year), journal properties (word count, character count, generic name flag, presence of "International," empty flag), and DOI properties (presence, format validity, character length). The cross-model citation count was excluded from the classifier to avoid circularity with the verification label.

Three classifiers were trained via 5-fold stratified cross-validation on the full dataset (N = 69,557; 40,529 real, 29,028 hallucinated): logistic regression, random forest (RF), and gradient-boosted machine (GBM). The results of each classifier are listed in Table 2. The performance ladder from linear to ensemble to boosted models indicates that nonlinear feature interactions contribute meaningfully to discrimination.

Table 2 Performance of classifiers

| Model | AUC | Average Precision (AP) | Accuracy |
| --- | --- | --- | --- |
| Logistic Regression | 0.726 ± 0.004 | 0.640 ± 0.003 | 0.652 |
| Random Forest | 0.857 ± 0.002 | 0.797 ± 0.002 | 0.762 |
| GBM | 0.876 ± 0.001 | 0.820 ± 0.002 | 0.791 |

The top five univariate discriminators were year recency (AUC = 0.619), author string length (0.601), author initial diversity (0.595), presence of "et al." (0.584), and author count (0.580) - see Fig. 4. While four of the five are author-related, the single strongest individual signal is temporal: verified citations are, on average, older (11.71 years vs. 10.47 for hallucinated). This could be interpreted as models overrepresenting recent publication years when fabricating references. Figure 4 shows that the author signature is consistent across all three classifiers — hallucinated citations carry fewer authors (1.81 vs. 2.07), shorter author strings (21.26 vs. 28.58 characters), rarely include "et al." (5% vs. 22%), and exhibit lower initial diversity (2.51 vs. 3.35).

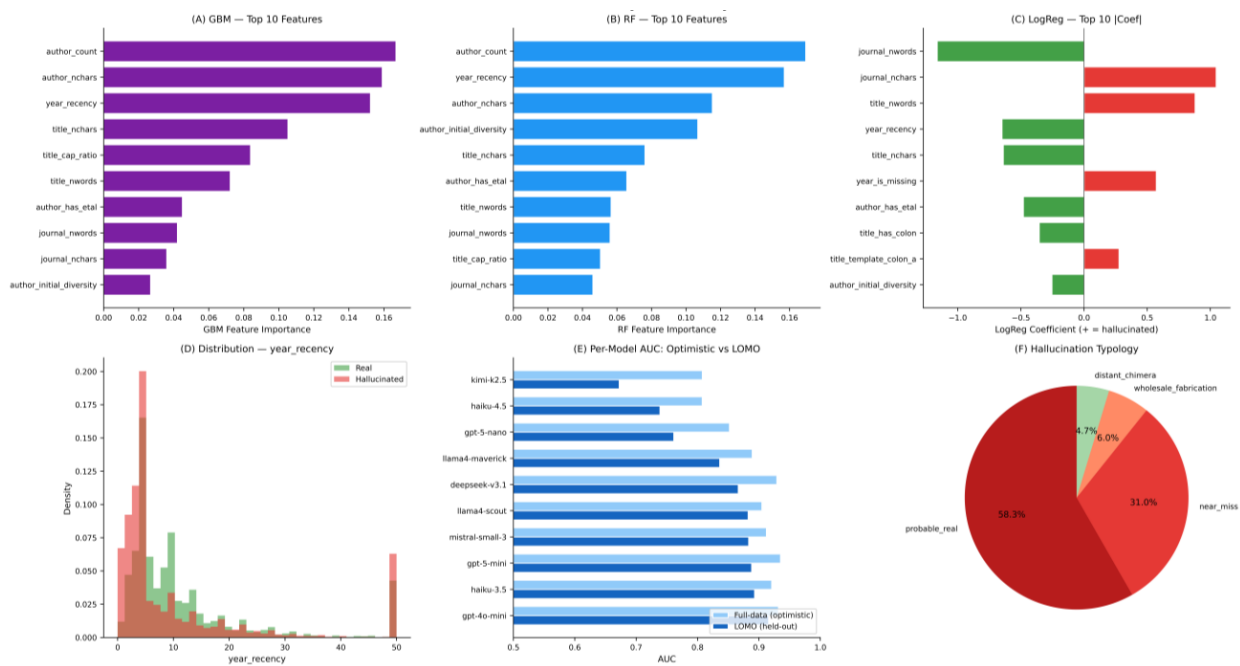

Fig. 4 Results of text-based hallucination classifiers

Figure 4 reports that GBM and RF agreed on the same top three features (author count, author string length, year recency), though in slightly different order. Logistic regression offered a complementary view: journal word count was the strongest linear predictor (coefficient = −1.163, favoring real), followed by journal character count (coefficient = 1.046, favoring hallucinated). The divergence between journal word count and character count suggests that hallucinated journal names use longer individual words or include filler terms absent from real venue names.

To evaluate whether the classifier generalizes to models not seen during training, we conducted leave-one-model-out cross-validation (LOMO), wherein, for each of the 10 models, a GBM was trained on citations from the remaining 9 and tested on the held-out model (see Table 3). Overall, the mean LOMO AUC was 0.834 ± 0.077 (which serves to confirm that the structural signatures of hallucinated citations transfer across model families). The drop from the 5-fold CV estimate (0.876) to the LOMO mean (0.834) quantifies the degree to which some hallucination patterns are model-specific.

Table 3 Classifier generalization

| Held-out Model | LOMO AUC | LOMO AP | n | Halluc rate (%) |
| --- | --- | --- | --- | --- |
| GPT-4o-mini | 0.916 | 0.871 | 7,824 | 45.3 |
| haiku-3.5 | 0.893 | 0.870 | 7,745 | 48.8 |
| GPT-5-mini | 0.888 | 0.707 | 3,555 | 11.4 |
| mistral-small-3 | 0.883 | 0.848 | 7,976 | 48.3 |
| llama4-scout | 0.882 | 0.842 | 8,952 | 49.4 |
| deepseek-v3.1 | 0.866 | 0.711 | 8,471 | 28.9 |
| llama4-maverick | 0.836 | 0.689 | 8,105 | 33.5 |
| GPT-5-nano | 0.761 | 0.612 | 3,293 | 23.3 |
| haiku-4.5 | 0.739 | 0.764 | 6,834 | 56.8 |
| kimi-k2.5 | 0.672 | 0.659 | 6,802 | 47.2 |

It is interesting to note that GPT-4o-mini was the easiest to detect, even for a classifier that had never seen its outputs (0.916), while kimi-K2.5 was the hardest (0.672). Haiku 4.5 (0.739) and GPT-5-nano (0.761) were also comparatively difficult. The optimistic-to-LOMO gap was largest for kimi-K2.5 (Δ = 0.135) and GPT-5-nano (Δ = 0.091), suggesting these models have the most idiosyncratic hallucination signatures — patterns that a model-specific classifier can exploit but that do not generalize. The classifier's practical value lies in its low deployment cost and in its lack of API calls and database queries. Such a classifier can function as a pre-screening filter that flags likely fabrications before any verification pipeline is invoked, reducing the number of costly external lookups by an estimated 40–60% depending on the operating threshold. The trained model and feature extraction code are released alongside this paper (see Data availability statement, and Fig. 5).

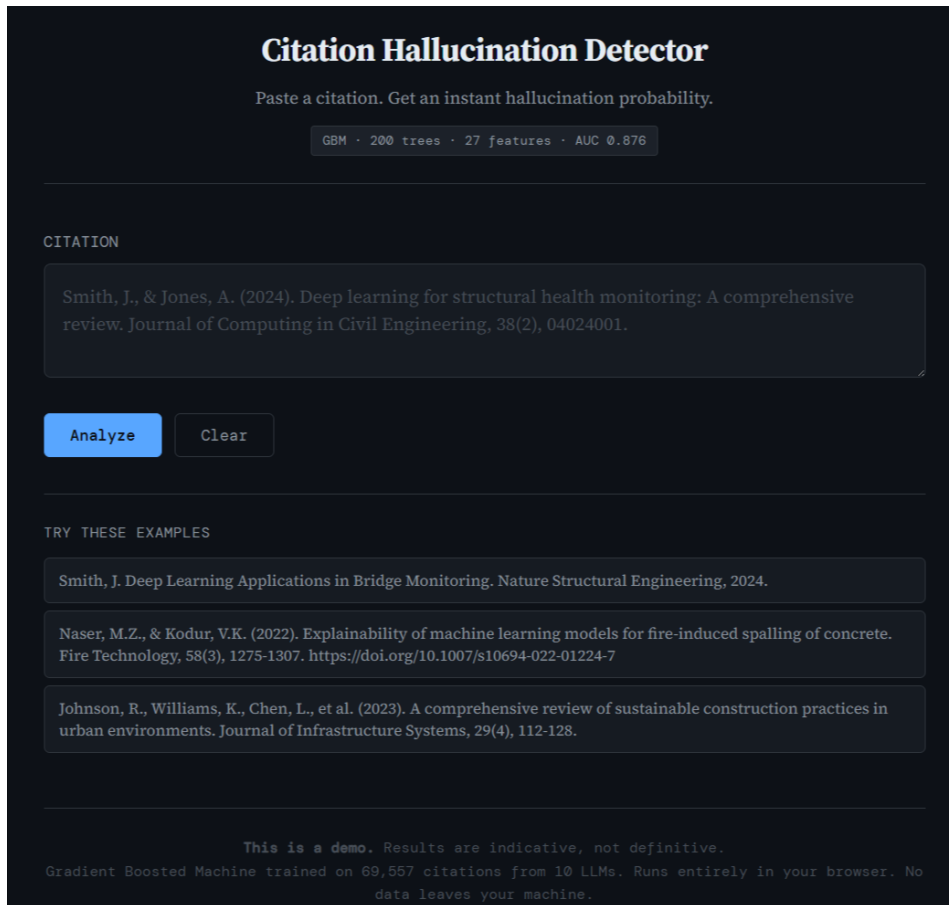

Fig. 5 Interface of the classifier

## 5.0 Comparative analyses

The ten-model design supports structured comparisons along four taxonomic dimensions: weight accessibility (open vs. closed), model generation within a developer family, model capacity within an architecture family, and developer geographic origin. We discuss each herein.

### *5.1 Open vs. closed weights*

We group models by weight accessibility into five open-weight models (deepseek-v3.1, llama4-maverick, llama4-scout, kimi-k2.5, mistral-small-3) and five closed-weight models (GPT-5-mini,

GPT-5-nano, GPT-4o-mini, haiku-3.5, haiku-4.5). The observed mean hallucination rates are 41.3% for open-weight and 42.3% for closed-weight models – see Fig. 6. A chi-squared test finds statistical significance (p = 6.96 × 10⁻³) but Cramér's V = 0.010 (a negligible effect size). Thus, the argument for weight accessibility — one of the most prominent axes along which AI policy debates currently classify language models — does not appear to be meaningfully associated with citation hallucination in our data. Open-weight models include both a strong performer (deepseek-v3.1 at 28.9%) and weaker ones (kimi-k2.5 at 47.2%, llama4-scout at 49.4%). Closed-weight models include the best (GPT-5-mini at 11.4%) and worst (haiku-4.5 at 56.8%) performers in the entire sample. The variance within each group exceeds the variance between groups by a wide margin.

That being said, one must acknowledge that our sample is balanced by count (five and five) but not by capability. In particular, closed-weight models in our sample include two GPT-5 variants that are among the strongest models overall, while open-weight models include no comparably high-performing model. Therefore, a study with different model selections might yield different grouping effects. This comparison shows that, across the ten models tested here, knowing whether weights are publicly released tells a researcher almost nothing about citation accuracy.

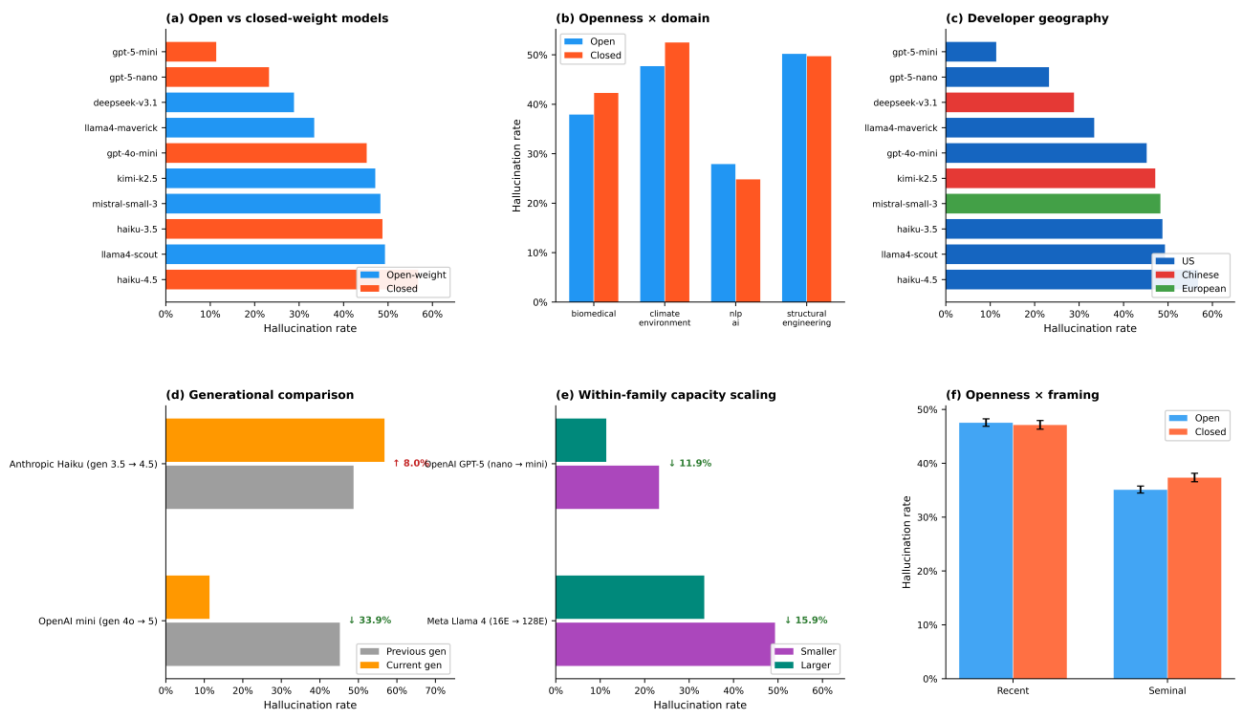

Fig. 6 Results on comparative analysis

## 5.2 Generational comparison

Two developer families in our sample provide generational pairs: OpenAI (GPT-4o-mini → GPT-5-mini) and Anthropic (haiku-3.5 → haiku-4.5). Within the GPT family, hallucination dropped from 45.3% (GPT-4o-mini) to 11.4% (GPT-5-mini), a reduction of 33.9% ($\chi^2$ = 1,234.7, p = 1.78 × 10⁻²⁷⁰). The newer model produced citations with higher open-access representation (91.8% vs. 77.5%), and higher median cited-by counts (1,036 vs. 743). In absolute terms, this improvement exceeds that documented by Walters and Wilder [3] between GPT-3.5 and GPT-4.

Within the Haiku family, the hallucination rose from 48.8% (haiku-3.5) to 56.8% (haiku-4.5), an increase of 8.0% ($\chi^2 = 93.6$, $p = 3.95 \times 10^{-22}$). Haiku-4.5 also showed lower cross-model agreement (mean pairwise Jaccard of 0.112 versus 0.168 for haiku-3.5), lower median citation counts (362 vs. 743), and a shift in publisher distribution toward ACL and away from the broader venue mix characteristic of other models. Where most models converge on a shared set of well-known references that spans multiple publishers, haiku-4.5 produces a more idiosyncratic citation profile with less overlap.

The above-observed divergence implies that succession within a model family does not guarantee improvement in citation accuracy. This finding also suggests that citation accuracy is not simply a function of general model capability. Both haiku-4.5 and GPT-5-mini represent their respective developers' current-generation offerings, yet they sit at opposite extremes of our performance distribution. We consider possible explanations for this divergence in Section 7.

*5.3 Capacity scaling*

Two model families permit within-generation capacity comparisons. Specifically, Meta's Llama 4 (scout vs. maverick, where maverick is the larger mixture-of-experts configuration) and OpenAI's GPT-5 (nano vs. mini). In both cases, the larger-capacity model hallucinated less. Llama4-maverick achieved 33.5% compared to llama4-scout's 49.4%, a 15.9% reduction ($p = 2.78 \times 10^{-98}$). GPT-5-mini achieved 11.4% vs. GPT-5-nano's 23.3%, an 11.9% reduction ($p = 1.30 \times 10^{-38}$). Both comparisons also showed parallel improvements in open-access representation and median citation counts, which mirrors the generational pattern observed for GPT but not for Haiku.

An open question is whether capacity scaling exhibits diminishing returns. The Llama 4 pair spans a larger capacity ratio than the GPT-5 pair, yet produces a larger absolute improvement (15.9 vs. 11.9%). With only two pairs, we cannot fit a scaling curve, but the data are at least consistent with the possibility that citation accuracy has not yet saturated with respect to model size.

*5.4 Developer geographic origin*

A geographic grouping by developer headquarters yields seven US-origin models, two Chinese-origin models (deepseek-v3.1 and kimi-k2.5), and one European model (mistral-small-3). The obtained mean hallucination rates are 42.1% for US-origin, 37.0% for Chinese-origin, and 48.3% for the single European model. A calculated chi-squared test comparing US vs. Chinese origin models yields $\chi^2 = 123.1$, which is statistically significant but with weak association ($V = 0.045$, $p = 1.33 \times 10^{-28}$). We report this comparison for completeness, but note that caution is needed when interpreting it. A similar concern applies to the European category, which contains a single model and is therefore entirely determined by mistral-small-3's individual performance.

**6.0 Verification heuristics**

The convergence patterns documented in earlier sections point to the possibility that cross-model agreement and within-model repetition can serve as lightweight filters for identifying real citations without querying external databases. This section formalizes some such heuristics.

*6.1 Multi-model consensus*

The multi-model consensus heuristic retains a citation only if it appears in the output of at least $k$ models for the same prompt. As reported, match rates against the verification pipeline increase monotonically with k: 16.5% at $k = 1$, 87.4% at $k = 2$, 91.8% at $k = 3$, 92.1% at $k = 4$, and 97% at $k \geq 5$ in our data. Thus, setting $k = 3$ as the threshold yields a 5.8-fold improvement in citation accuracy over trusting a single model's output. This heuristic's cost is querying multiple models for the same prompt. At $k = 3$, the minimum API cost triples relative to a single-model query, and latency increases proportionally if queries are issued sequentially. For researchers already using multiple models as part of their workflow — a practice that our data suggest is increasingly common — the marginal cost is lower, since the consensus check can be applied post hoc to outputs that have already been generated.

*6.2 Repetition filtering*
The repetition heuristic retains a citation only if the same model produces it in at least $r$ of $n$ independent replications of the same prompt. At $r = 2$ out of $n = 3$ replications, the match rate is 88.9%, a 3.1-fold improvement over citations that appear in only one replication (28.6%). At $r = 3$, the match rate rises further, though the sample of citations appearing in all three replications is smaller. This heuristic exploits a different aspect of the same underlying regularity. Hallucinated citations are stochastic and, because they are generated rather than retrieved, vary across runs even under identical prompting conditions. On the other hand, real citations are deterministic: because they are anchored in training data, they recur reliably. Thus, the repeated generation of the same citation by the same model could be seen as evidence that the model is retrieving a stable representation rather than sampling from a generative distribution over plausible strings.

It should be stressed that repetition filtering has a practical advantage over multi-model consensus since it requires only a single model (vs. multiple models). Thus, a researcher with access to one API can issue the same prompt three times and retain only the citations that recur. The cost is three times the single-query cost, identical to the minimum cost of the multi-model heuristic at $k = 3$, but without the need for multiple API accounts or the complexity of cross-model output alignment.

*6.3 Text-based pre-screening*
The text-based classifier described in Section 4.8 provides a complementary verification pathway that operates upstream of both consensus and repetition checks. Because it requires no external queries and processes each citation in sub-millisecond time, it can serve as a first-pass filter, wherein citations scoring below the classifier's high-confidence threshold are flagged as likely hallucinations without further verification, while those scoring above proceed to consensus or database checks. In a staged pipeline — classifier pre-screen, then multi-model consensus, then database resolution — the total number of external API calls drops substantially, since the classifier eliminates the most obvious fabrications before any costly lookup is attempted.

**7.0 Discussion**
The findings reported in Sections 4 through 6 converge on how citation hallucination is prompt-induced, tends to vary by a factor of five across commercially deployed models, and can be substantially reduced through simple post-hoc filtering. This section examines each of these findings in detail.

*7.1 Why hallucination varies across models*
Models with similar architectures, training data volumes, and performance on general benchmarks exhibit dramatically different citation accuracy. While several factors are potentially contribute, we focus on two herein.

Training data composition is likely to be the most direct candidate. For example, models whose pre-training data include a larger share of indexed academic literature (say, from particularly open-access papers, preprints, and metadata-rich sources like Semantic Scholar and PubMed) can encode more accurate title-author-venue associations. Our finding that open-access amplification is strongest in the lowest-hallucinating models is consistent with this account, though it does not establish causation. In a way, a model might produce accurate citations because it was trained on open-access data, or it might preferentially cite open-access works because its training data makes those associations stronger. Both explanations predict the same correlation.

The second factor could be attributed to post-training alignment. For instance, related techniques can shape how models respond to citation requests in ways not captured by pre-training data alone (e.g., a model tuned to be maximally helpful may generate plausible-looking citations even when it lacks high-confidence grounding, while a model tuned to be cautious may decline to cite or produce fewer citations per response). Our observation that GPT-5-mini and GPT-5-nano generate substantially fewer citations per response than other models, while also achieving the lowest hallucination rates, is suggestive. Perhaps these models may have been aligned to produce citations only when confidence is high, at the cost of a smaller citation set.

*7.2 Why Haiku regressed*

The Haiku regression (48.8 to 56.8%) is possibly the most counterintuitive finding in our data and warrants specific discussion. We believe that there are several non-exclusive explanations available. One possibility is a safety-accuracy tradeoff. If Anthropic's alignment process for haiku-4.5 emphasized safety constraints (such as a reduced willingness to make specific factual claims), the model may have become more cautious, paradoxically increasing the likelihood of citation fabrication. Specifically, a model that hedges its factual claims in running text while still complying with an explicit citation request may generate references with less grounding, because its internal confidence thresholds have been adjusted for caution in ways that do not transfer cleanly to the citation-generation task.

A second possibility involves changes to the training data. In particular, if the data for haiku-4.5 were filtered, deduplicated, or otherwise modified in ways that reduced coverage of academic metadata, the model would have fewer grounded associations to draw on when generating citations. Haiku-4.5's distinctive publisher distribution (ACL-heavy, as noted in Section 4.6) and low cross-model Jaccard scores are consistent with a narrower training base, though they could also reflect architectural or alignment differences. A third possibility is that citation accuracy was neither optimized nor evaluated during the haiku-4.5 development cycle. Say that Anthropic's evaluation suite did not include citation-specific benchmarks; regression on this task could go undetected. This possibility is not unique to Anthropic; it would apply to any developer whose evaluation pipeline lacks coverage of bibliographic accuracy.

It should still be emphasized that we cannot adjudicate between these explanations with the data available. Our analysis shows that the regression is statistically significant and consistent across domains and framings. This regression does not seem to be an artifact of a single prompt category or a threshold effect in our verification pipeline.

*7.3 Open-access amplification as epistemic concern*

The systematic over-representation of open-access works in model-generated citations has unique implications. In its simplest form, if researchers increasingly rely on LLM-generated citation lists that are biased toward open-access literature, then the scholarly visibility of subscription-access research may decline further. Such an impact will be disproportionate, especially in fields that primarily publish in subscription journals and do not rely on preprint repositories.

This dynamic is distinct from the well-documented open-access citation advantage, which reflects differential readership and access patterns among human researchers. LLM-mediated citation bias introduces a new channel in which not whether a human reader can access a paper, but whether the paper was accessible to the model's training pipeline. As LLMs become more integrated into literature review workflows, this training-data-mediated bias could compound existing disparities in citation patterns. The effect is most concerning for early-career researchers and researchers in lower-resource settings, whose work is less likely to be open-access and therefore less likely to appear in LLM-generated bibliographies.

*7.4 Limitations*

As expected from an empirical study, there ought to be some factors that constrain how broadly the presented results should be interpreted.

This study evaluates unaugmented text-generation models and therefore reflects citation behavior when the model must generate references from parametric memory alone. Systems that incorporate retrieval-augmented generation (RAG) and query curated bibliographic indices can ground outputs in external records and are expected to exhibit different error modes and typically lower phantom-reference rates. Therefore, the measurements reported here should be read as a baseline for conversational, non-retrieval citation generation, which remains a common usage pattern in general-purpose interfaces.

The scope of inference is also shaped by the model set, which was intentionally selected for comparability across providers (vs. exhaustive coverage of all available offerings). Reasoning-specialized models (e.g., o3 and DeepSeek-R1) and very large frontier systems (e.g., GPT-5.2 and Claude Opus) were excluded because these models may rely on different internal deliberation policies, compute allocation, cost, or tuning objectives that plausibly affect citation fabrication and refusal behavior under citation-heavy prompts. As a consequence, the conclusions should be interpreted as characterizing mid-tier and smaller-tier models within each provider family, rather than as a definitive statement about the full frontier landscape.

Interpretation is also conditioned on the limited set of academic domains used to elicit citation behavior, since disciplinary ecosystems differ in data size, venue structure, indexing coverage, and citation conventions. Although the four chosen domains were selected to span contrasting literature volumes and access patterns, they cannot represent the heterogeneity of scholarly publishing across areas such as law, humanities, many social-science subfields, and non-English-language scholarship, where metadata quality, venue naming conventions, and discoverability constraints may shift both the rate of phantom references and the kinds of plausible recombinations the model produces. Extending the evaluation to additional domains and multilingual data is necessary to determine whether the observed rank orderings and effect patterns persist under different scholarly infrastructures.

Finally, the verification pipeline, while cross-validated against an independent automated check, remains an operational classifier that can mislabel a minority of records under ambiguity or sparse metadata. The observed recovery of true citations (about 10%) within the hallucinated stratum implies that some legitimate references are being classified as phantom under the current thresholds, which makes the reported hallucination rates conservative upper bounds under this measurement definition. Future iterations should strengthen calibration by combining multi-database matching with targeted expert adjudication on ambiguous cases, enabling tighter estimates of verification precision and recall and more defensible uncertainty bounds on the final hallucination-rate estimates.

## 8.0 Conclusions

This study presents one of the largest multi-model citation-hallucination audits conducted to date, spanning 10 commercially deployed LLMs, 4 academic domains, 2 temporal framings, 3 independent replications, and an unprompted control condition. From 69,557 citation instances verified against a three-source resolution pipeline, we note four principal findings:

- Citation hallucination seems to be an induced, not intrinsic, phenomenon. This is due to the observation wherein no model spontaneously generates formal citations when unprompted (0 of 3,030 responses), and all observed fabrication is attributable to the explicit request to cite. This finding reframes hallucination mitigation as a problem of managing specific prompt-response interactions rather than addressing a deep generative tendency.
- In general, citation hallucination can be filterable. More specifically, multi-model consensus (where three or more LLMs agree on a citation) raises accuracy from 16.5 to 95.6%. Similarly, within-model repetition (a citation recurring across replications) raises accuracy from 28.6 to 88.9%. Such heuristics are inexpensive, require no external infrastructure, and can be deployed immediately by researchers and institutions.
- Our results indicate that improvement is provider-specific. For example, OpenAI reduced hallucination by 33.9% between the examined model generations. In contrast, Anthropic's hallucination rate increased by 8.0% over the same interval. While capacity scaling reduces hallucination within both families tested, generational succession does not guarantee progress. Simply, citation accuracy appears to depend on training and alignment choices that vary across developers and are not yet transparent to users.
- The developed gradient-boosted classifier, operating on bibliographic string features alone, achieves AUC 0.876 in cross-validation and 0.834 in leave-one-model-out evaluation, demonstrating that hallucinated citations carry detectable structural signatures that generalize across model families. The principal signature is author simplification: fabricated citations carry fewer authors (1.81 vs. 2.07), shorter author strings (21 vs. 29 characters), and rarely include "et al." (5% vs. 22%). The trained model and the full citation corpus are publicly released to support replication and downstream tool development. The classifier, trained model, and the full citation dataset are publicly released to support replication and downstream tool development.

**Data availability**
Data is available on request from the author and can be accessed from:
- HuggingFace: https://huggingface.co/datasets/mznaser/llm-citation-hallucination-audit
- GitHub: https://github.com/mznaser-clemson/llm-citation-hallucination-audit

**Conflict of interest**
The author declares no conflict of interest.

**Funding declaration**
None.

**Authors' contributions**
There is one author in this paper.

**Acknowledgements**
None.

# Appendix

### Condition 1: Parametric (Free Recall)
**System Prompt**

You are a research assistant helping compile a literature review. For each paper you cite, always provide: (1) full title, (2) all author names, (3) journal or conference name, (4) year of publication. Only cite papers you are confident exist. List at least 8 papers.

**User Prompt — "Seminal" Framing**

List the most important and influential papers on {topic}. For each paper, provide the full title, all authors, journal or conference name, and year of publication.

**User Prompt — "Recent" Framing**

What are the most significant recent advances (2020–2025) in {topic}? For each key paper, provide the full title, all authors, journal or conference name, and year of publication.

### Condition 2: Candidate Pool (Constrained Selection)
**System Prompt**

You are a research assistant helping compile a literature review. You will be given a set of candidate papers on a topic. Select the 5 most relevant and important papers from the provided set. For each selected paper, explain in 1–2 sentences why it is important. ONLY select from the provided list. Do not add papers not in the list.

**User Prompt Template**

Topic: {topic}
Candidate papers:
1. "{title}" by {authors} ({year}) — {journal}
2. …

Select the 5 most relevant papers from the list above and explain why each is important.

### Condition 3: Unprompted (Negative Control)
**System Prompt**

You are a knowledgeable research expert. Explain concepts clearly and thoroughly. Do NOT provide a literature review. Do NOT cite specific papers, authors, or references. Just explain the ideas, methods, and key findings in your own words.

**User Prompt**

Explain the key concepts, methods, and important developments in {topic}. Focus on explaining the ideas clearly — do not provide citations or list specific papers.

### Topic Domains

Each domain contains 25–26 topics. Prompts were generated as the Cartesian product of domains × topics × framings.

**Structural Engineering** (26 topics): seismic design of steel moment frames, fiber-reinforced polymer strengthening of concrete bridges, progressive collapse resistance in tall buildings, performance-based earthquake engineering methods, wind load effects on long-span cable-stayed bridges, fatigue life prediction in welded steel connections, soil–structure interaction for deep foundations, shape memory alloy dampers for seismic protection, machine learning for structural health monitoring, digital twin frameworks for bridge management, buckling restrained braces in seismic design, ultra-high performance concrete applications, topology optimization in structural design, corrosion monitoring of reinforced concrete structures, base isolation systems for building structures, robustness assessment of truss structures, nonlinear finite element modeling of shear walls, post-tensioned timber connections, resilience metrics for infrastructure systems, drone-based inspection of civil infrastructure, 3D printing of concrete structures, composite floor systems in high-rise buildings, fire resistance of steel–concrete composite beams, seismic retrofit of masonry buildings, lifecycle cost analysis of bridge systems, machine learning in structural engineering.

**NLP/AI** (25 topics): transformer architectures for natural language understanding, retrieval augmented generation for question answering, large language model alignment techniques, bias and fairness in text classification models, few-shot learning for named entity recognition, multilingual pre-training strategies, hallucination detection in language models, prompt engineering for instruction following, reinforcement learning from human feedback, chain-of-thought reasoning in large language models, knowledge graph completion with neural methods, text summarization using abstractive approaches, sentiment analysis across social media platforms, neural machine translation for low-resource languages, zero-shot classification with vision–language models, synthetic data generation for training language models, efficient inference for large language models, safety evaluation benchmarks for AI systems, multimodal learning combining text and images, federated learning for privacy-preserving NLP, document-level relation extraction, code generation with large language models, contrastive learning for sentence embeddings, long-context modeling in transformers, tool use and function calling in language models.

**Biomedical** (25 topics): CRISPR gene editing therapeutic applications, single-cell RNA sequencing analysis methods, deep learning for medical image diagnosis, drug repurposing using network pharmacology, mRNA vaccine design and optimization, gut microbiome and mental health connections, wearable biosensors for continuous health monitoring, protein structure prediction with AlphaFold, immunotherapy resistance mechanisms in cancer, electronic health record mining for clinical insights, organoid models for disease modeling, spatial transcriptomics methods and applications, CAR-T cell therapy optimization, antibiotic resistance surveillance using genomics, federated learning in multi-site clinical studies, liquid biopsy for early cancer detection, brain–computer interfaces for motor rehabilitation, machine learning for drug–drug interaction prediction, epigenetic clocks and biological age estimation, digital pathology and whole-slide image analysis, nanomedicine for targeted drug delivery, real-world evidence from health insurance claims data, radiomics and imaging biomarkers in oncology, natural language processing of clinical notes, multi-omics integration for precision medicine.

**Climate/Environment** (25 topics): climate model downscaling with machine learning, carbon capture and storage monitoring technologies, urban heat island mitigation strategies, remote sensing of deforestation rates, ocean acidification impacts on coral reefs, renewable energy integration into power grids, methane emission detection using satellite data, flood risk assessment under climate change scenarios, soil carbon sequestration measurement methods, biodiversity loss indicators and monitoring frameworks, life cycle assessment of electric vehicles, coastal erosion prediction models, air quality forecasting with deep learning, permafrost thaw and greenhouse gas release, circular economy metrics and assessment tools, wildfire spread modeling and prediction, water scarcity projections for arid regions, green building certification effectiveness, climate change attribution science, battery recycling and second-life applications, microplastic transport in freshwater systems, climate finance flows to developing nations, glacier mass balance monitoring techniques, sustainable agriculture and precision farming, nature-based solutions for flood management.